\documentclass[runningheads]{llncs}
\usepackage{graphicx}
\usepackage{adjustbox}
\usepackage{comment}
\usepackage{amsmath,amssymb} %
\usepackage{color}
\usepackage[abs]{overpic}

\usepackage{multirow}
\usepackage{tabulary}
\usepackage{makecell}
\usepackage{tabularx,ragged2e}
\usepackage{booktabs}
\usepackage[font=small,labelfont=bf]{caption}

\newcommand{\keypoint}[1]{\vspace{0.1cm}\noindent\textbf{#1}\quad}
\newcommand{\cut}[1]{}

\newcommand{\fullname}{What-Where-When}%
\newcommand{\shortname}{W3}%

\begin{document}
\pagestyle{headings}
\mainmatter
\def\ECCVSubNumber{1052}  %

\title{Knowing What, Where and When to Look: Efficient Video Action Modeling with Attention}

\titlerunning{Knowing What, Where and When to Look}
\authorrunning{P\'erez-R\'ua et al.}
\author{Juan-Manuel P\'erez-R\'ua\inst{1} \and
Brais Martinez\inst{1} \and Xiatian Zhu\inst{1} \and \newline
Antoine Toisoul\inst{1} \and Victor Escorcia\inst{1} \and Tao Xiang\inst{1,2}
}
\institute{Samsung AI Centre, Cambridge, UK \email{\{j.perez-rua,brais.a,xiatian.zhu,a.toisoul,v.castillo,tao.xiang\}@samsung.com} \and
University of Surrey, UK}

\maketitle

\begin{abstract}
Attentive video modeling is essential for action recognition in unconstrained videos %
due to their rich yet redundant information over space and time.  
However, introducing attention in a deep neural network for action recognition is challenging for two reasons. 
First, an effective attention module needs to learn {\em what} (objects and their local motion patterns),
{\em where} (spatially), and {\em when} (temporally) to focus on. 
Second, a video attention module must be efficient because existing action recognition models already suffer from high computational cost. 
To address both challenges, a novel What-Where-When (W3) video attention module is proposed. 
Departing from existing alternatives, our W3 module models all three facets of video attention jointly. 
Crucially, it is extremely efficient by
factorizing the high-dimensional video feature data into low-dimensional meaningful spaces 
(1D channel vector for `what' and 2D spatial tensors for `where'), 
followed by lightweight temporal attention reasoning. 
Extensive experiments show that our attention model brings significant improvements to existing action recognition models, 
achieving new state-of-the-art performance on a number of benchmarks.


\end{abstract}

\section{Introduction}

Human action recognition in unconstrained videos remains 
an unsolved problem, particularly as the recent research interest has shifted to fine-grained action categories involving interactions between humans and objects \cite{damen2018scaling,goyal2017something,mahdisoltani2018fine,zhang2018egogesture}.  Why is a simple action such as ``{\em placing something next to something}'' so hard for a computer vision system yet so trivial for humans? One explanation is that 
videos typically contain highly redundant information in space and time which distracts a vision model from computing discriminative representations for recognition. For instance, with a cluttered background, there could be many other objects in the scene which can also interact with humans. Removing such distracting information from video modeling poses a great challenge.  Human vision systems, on the other hand, have highly effective attention mechanisms to focus on the most relevant objects and motion patterns (what) at the right place (where) and time (when) \cite{Rensink2000humanAttention}. Introducing attention modeling in a video action recognition model is therefore not only useful but also essential. 

Although attention modeling has been universally adopted in recent natural language processing (NLP) studies \cite{bahdanau2014neural,devlin-etal-2019-bert,gehring2017convolutional,luong2015effective,vaswani2017attention},
and many static image based computer vision problems
\cite{fu2019dual,gu2018learning,huang2019ccnet,park2018bam,woo2018cbam} in the deep learning era, it is much understudied in action recognition. This is due to a fundamental difference: there are two facets in attention modeling for NLP (what and when), as well as static image (what and where), but three for video (what, where and when). Adding one more facet into attention modeling brings about significant challenges in model architecture design, training and inference efficiency. As a result, existing attentive action recognition models \cite{du2017recurrent,li2018videolstm,meng2019interpretable,wang2018non,wang2016hierarchical} only focus on a subset of the three facets. Among them, only the self-attention based non-local module \cite{wang2018non} shows convincing benefits and is  adopted by recent 3D CNN-based models. However, it adds significant computational cost (see Table \ref{tab:att}) and is known to be hard to train \cite{Tao2018nonlocalproblem}.

In this paper, a novel {\em \fullname{}} (\shortname{}) video attention module is proposed. As suggested by the name, it models all three facets of video attentions. Crucially, it is extremely efficient, only adding marginal computational overhead in terms of floating point operations when 
introduced in an existing action recognition model. 
Inspired by the recent spatial and channel-wise factorization adopted in static image attention \cite{woo2018cbam}, 
the key idea of \shortname{} is to %
factorize channel and spatial attention into static and temporal components.
More concretely, given a feature map from a video CNN block/layer denoted as a tensor of four dimensions (time, channel and two for space), two attention sub-modules are formulated. 
In the first sub-module, each channel (representing object categories and  local movement patterns, i.e., `what') is attended, followed by 1D CNN based temporal reasoning. 
It is thus designed to capture important local motions evolving over time. In the second sub-module, spatial attention evolution over time is modeled using a lightweight 3D CNN to focus on the `where' and `when' facets. 
The two sub-modules are then applied to the original feature map sequentially and trained end-to-end with the rest of the model.  
Even after extensive factorization, the introduction of temporal reasoning in both sub-modules makes our \shortname{} hard to train. 
This is due to the vanishing gradient problem when our attention module with temporal reasoning is added to each layer of a state-of-the-art deep video CNN models such as I3D \cite{carreira2017quo} or TSM \cite{lin2019tsm}. 
To overcome this problem, two measures are taken. 
First,  gradients are  directed to the attention modules of each layer by an attention guided feature refinement module. 
Second, Mature Feature-guided Regularization (MFR) is formulated based on staged self-knowledge-distillation to prevent the model from being stuck in a bad local minimum.

The main {\bf contributions} of this work are as follows. {\bf (1)}  We introduce a novel \fullname{} (\shortname{}) video attention module
for action recognition in unconstrained videos. It differs from existing video attention modules in that all three facets of attention are modeled jointly. {\bf (2)} The computational overhead of the proposed attention module is controlled to be marginal (e.g., merely $1.5\%\sim3.2\%$ extra cost on TSM) thanks to a new factorized network architectural design for video attention modeling. {\bf (3)} The problem of effective training of a deep video attention module is addressed with a novel combination of attention guided feature refinement module and mature feature-guided (MFR) regularization. 
Extensive experiments are conducted on four fine-grained video action benchmarks including Something-Something V1 \cite{goyal2017something} and V2 \cite{mahdisoltani2018fine},
EgoGesture \cite{zhang2018egogesture}, and EPIC-Kitchens \cite{damen2018scaling}. The results show that our module brings significant improvements to existing action recognition models, achieving new state-of-the-art performance on a number of benchmarks.

\section{Related Work}

\keypoint{Video action recognition}
Video action recognition has made significant advances
in recent years, due to the availability of more powerful computing facilities (e.g., GPU and TPU),
the introduction of ever larger video datasets \cite{carreira2017quo,damen2018scaling,goyal2017something,karpathy2014large},
and the active developments of deep neural network based  action models \cite{feichtenhofer2019slowfast,lin2019tsm,Martinez_2019_ICCV,tran2018closer,wang2016temporal,wang2018non,wang2018videos,zhou2018temporal,zolfaghari2018eco}.
Early efforts on deep action recognition were focused on combining 2D CNN for image feature computing with  RNN for temporal reasoning
\cite{baccouche2011sequential,donahue2015long,yue2015beyond,zhou2018temporal} or 2D CNN on optical flow \cite{simonyan2014two}.
These have been gradually replaced by 3D convolutions (C3D) networks \cite{ji20123d,tran2015learning} recently.
The 3D kernels can be also formed via
inflating 2D kernels \cite{carreira2017quo} 
which facilitates
model pre-training using larger scale image datasets,
e.g., ImageNet. 

Two recent trends in action recognition are worth mentioning. First, the interest has shifted from coarse-grained categories such as those in UCF101 \cite{soomro2012ucf101} or Kinetics \cite{carreira2017quo} where background (e.g., a swimming pool for diving) plays an important role, to fine-grained categories such as those in Something-Something \cite{goyal2017something} and EPIC-Kitchens \cite{damen2018scaling} where modeling human-object interaction is the key. 
Second, since 3D CNNs typically are much larger and require much more operations during inference, many recent works focus on efficient network designed based on 2D spatial + 1D temporal factorization (R2+1D) \cite{MoreIsLess19,qiu2017learning,tran2018closer} or 2D+temporal shift module (TSM) \cite{Tran2019channel-separated}. TSM is particularly attractive because it has the same model complexity as 2D CNN and yet can still capture temporal information in video effectively. In this work we focus on fine-grained action recognition for which attention modeling is crucial and using TSM as the main backbone instantiation even though it can be applied to any other video CNN models.

\keypoint{Attention in action recognition}
Most existing video attention modules are designed for the out-of-date RNN based  action recognition models.  They are based on  either
encoder-decoder attention
\cite{du2017recurrent,li2018videolstm,meng2019interpretable,torabi2017action,wang2018action,wang2016hierarchical},
spatial attention only \cite{girdhar2017attentional,sharma2016action,wang2016hierarchical},
temporal attention \cite{torabi2017action,wang2018action},
or spatio-temporal attention \cite{du2017recurrent,li2018videolstm,meng2019interpretable,wang2016hierarchical}.
Compared to our \shortname{}, they are much weaker on the `what' facet -- our module attends to each CNN channel representing a combination of object and its local motion pattern only  when it evolves over time in a certain way. Further,   as they are formulated in the context of recurrent models, they cannot be integrated to 
the latest video CNN-based state-of-the-art action models. 
Other video attention methods are designed specifically 
for egocentric videos \cite{li2018eye,sudhakaran2018attention}
or skeleton videos \cite{si2019attention,song2017end,yang2018action}, 
which however assume specific domain knowledge (e.g., central objects \cite{sudhakaran2018attention}),
or extra supervision information (e.g., eye gaze \cite{li2018eye}),
or clean input data (e.g., skeleton \cite{si2019attention,song2017end,yang2018action}).
In contrast, our module is suited for  action understanding
in unconstrained videos without extra assumptions and supervision.

Note that the aforementioned video attention modules as well as our \shortname{} are non-exhaustive, 
focusing on a limited subset of the input space to compute attention. %
Recently, inspired by the success of transformer self-attention in NLP \cite{devlin-etal-2019-bert,vaswani2017attention}, non-local networks have been proposed \cite{wang2018non} and adopted widely \cite{girdhar2019video,lin2019tsm,Tran2019channel-separated}. 
By computing exhaustively the pairwise relationships between a given position and all others in space and time, 
non-local self-attention can be considered as a more generic and potentially more powerful attention mechanism than ours. 
However, a number of factors make it less attractive than \shortname{}. 
(1) Self-attention in NLP models use positioning encoding to keep the temporal information. When applied to video, the
non-local operator does not process any temporal ordering information (i.e., missing `when'), while temporal reasoning is performed explicitly in our attention module. 
(2) The non-local operator induces larger computational overhead  (see Table \ref{tab:att}) due to exhaustive pairwise relationship modeling and is known to have  convergence problems during training \cite{Tao2018nonlocalproblem}. 
In contrast, our \shortname{} adds neglectable overhead (see Table \ref{tab:att}), and is easy to train thanks to our architecture and training strategy specifically designed to assist in gradient flow during training. 
Importantly, our model is clearly superior to non-local when applied to the same action model (see Sec.~\ref{sec:experiments}).

\section{ What-Where-When Video Attention}

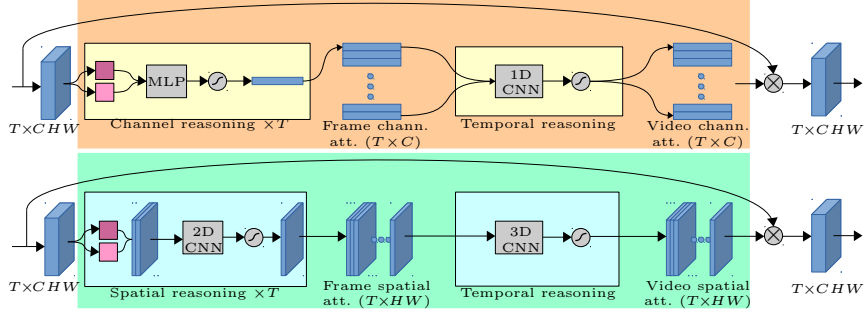
\begin{figure*}[t]
    \centering
    \begin{overpic}[width=0.95\textwidth]{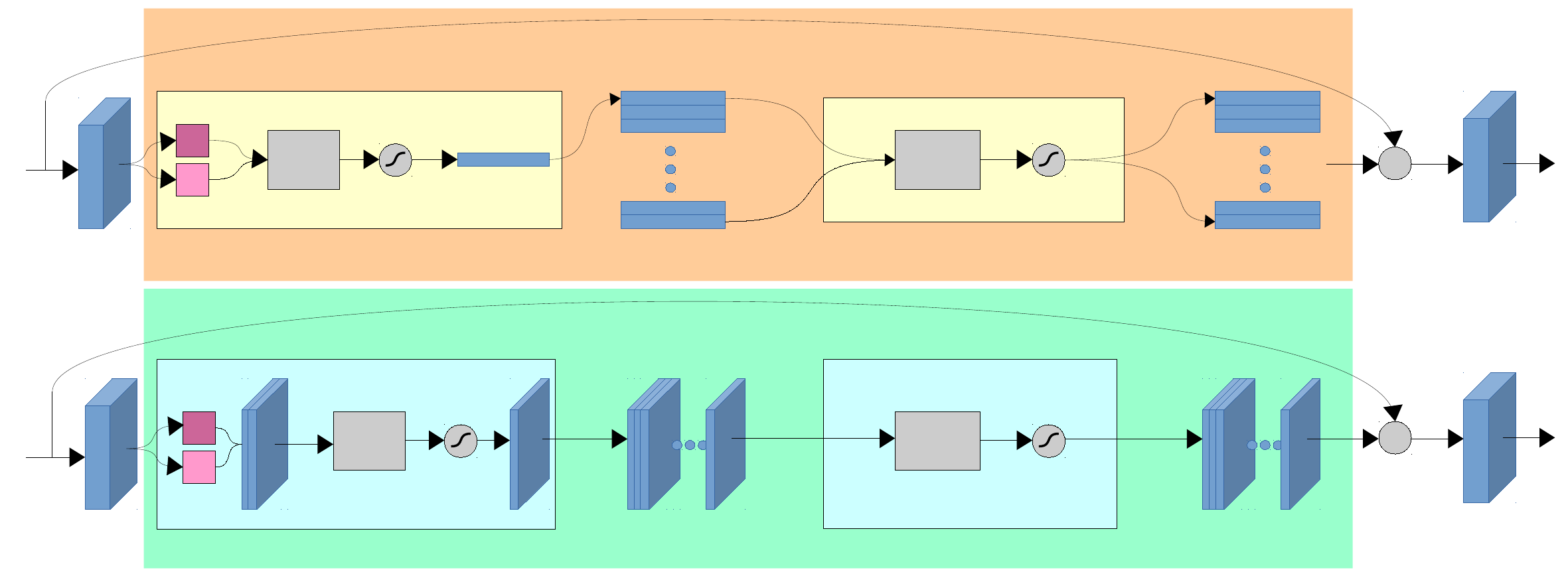}

    \put (4.0, 70.5) {\tiny{$T\hspace{-0.75mm}\times \hspace{-0.9mm} CHW$}}
    \put (4.0, 10.5) {\tiny{$T\hspace{-0.75mm}\times \hspace{-0.9mm} CHW$}}

    \put (44.0,  87.5) {\parbox{40pt}{\centering \tiny{MLP}}}
    \put (188.0,  87.5) {\parbox{20pt}{\centering \tiny{1D CNN}}}

    \put (42.5, 72.5) {\tiny{Channel reasoning $\times T$}}
    \put (175.0, 72.5) {\tiny{Temporal reasoning}}    

    \put (42.5, 8.5) {\tiny{Spatial reasoning $\times T$}}
    \put (175.0, 8.5) {\tiny{Temporal reasoning}}
    
    \put (68.0,  28.0) {\parbox{20pt}{\centering \tiny{2D CNN}}}
    \put (188.0,  28.0) {\parbox{20pt}{\centering \tiny{3D CNN}}}    
    
    \put (118.0,  67.5) {\parbox{50pt}{\centering \tiny{Frame chann. att. $(T\hspace{-0.5mm}\times\hspace{-0.5mm}C)$}}}
    \put (240.0,  67.5) {\parbox{50pt}{\centering \tiny{Video chann. att. $(T\hspace{-0.5mm}\times\hspace{-0.5mm} C)$}}}
    
    \put (240.0,  7.0) {\parbox{50pt}{\centering \tiny{Video spatial att. $(T\hspace{-0.75mm}\times\hspace{-0.9mm}HW)$}}}
    \put (118.5,  7.0) {\parbox{50pt}{\centering \tiny{Frame spatial att. $(T\hspace{-0.75mm}\times\hspace{-0.9mm}HW)$}}}
    
    \put (290.0,87.0) {\footnotesize{$\times$}}
    \put (290.0,29.5) {\footnotesize{$\times$}}    
    
    \put (300.0, 70.0) {\tiny{$T\hspace{-0.75mm}\times \hspace{-0.75mm} CHW$}}
    \put (300.0, 10.0) {\tiny{$T\hspace{-0.75mm}\times \hspace{-0.75mm} CHW$}}    
    
    \end{overpic}
    \caption{An overview of the proposed \shortname{} attention module.  {\bf Top:} Detail of the channel-temporal attention sub-module (orange box). A multi-layer perceptron transforms the 
    input feature into a per-frame attention vector. The concatenation of these vectors across the temporal dimension
    is further processed by a temporal CNN (1D convolutions) and a final sigmoid non-linearity.
    {\bf Bottom:} Detail of the spatio-temporal attention sub-module (green box). After a 2D convolution on the concatenation of cross-channel max and mean pooled features,
    a 3D CNN is applied on the stacked single-channel per-frame intermediate spatial attention maps. %
    Attention maps are point-wise multiplied with the input features. 
    For both blocks, the dark and light  purple boxes are \emph{max} and \emph{mean} pooling operations, respectively.}
    \label{fig:W3_detail}
\end{figure*}

\keypoint{Overview.}
Given an action recognition network based on a 3D CNN or its various lightweight variants, our \shortname{} is illustrated in Fig. \ref{fig:W3_detail}. We take a 4D feature map ${\bf F} \in \mathbb{R}^{T\times C\times H \times W}$ from any intermediate layer as the input of \shortname{}, 
where $T, C, H, W$ denote the frame number of the input video clip, the channel number, the height and width of the frame-level feature map respectively. 
Note that the feature map of each channel is obtained using a 3D convolution filter or a time-aware 2D convolution in the case of TSM networks~\cite{zhou2018temporal}; it thus captures the combined information about both object category and its local movement patterns, i.e., `what'. 
The objective of \shortname{} is to compute a same-shape attention mask ${\bf M} \in \mathbb{R}^{T\times C\times H \times W}$ that can be used to refine the feature map in a way such that action class-discriminative cues can be sufficiently focused on, whilst the irrelevant ones are suppressed.
Formally, this attention learning process is expressed as:
\begin{equation}
    {\bf F}' = {\bf F} \otimes {\bf M}, \;\;\;\; {\bf M} = f({\bf F})
\end{equation}
where $\otimes$ specifies the element-wise multiplication operation, and $f()$ is the W3 attention reasoning function. 

To facilitate effective and efficient attention learning, 
we consider an attention factorization scheme
by splitting the 4D attention tensor ${\bf M}$ into
a channel-temporal attention sub-module ${\bf M}^c \in \mathbb{R}^{T\times C}$
and 
a spatio-temporal attention sub-module ${\bf M}^s \in \mathbb{R}^{T\times H \times W}$.
This reduces the attention mask size from $TCHW$ to $T(C+HW)$ and therefore
the learning difficulty.
As such, the above feature attending is reformulated into a two-step sequential process as:
\begin{equation}
    {\bf F}^c = {\bf M}^c \otimes {\bf F}(T,C), \; {\bf M}^c = f^c({\bf F}); \;\;\;\;
    {\bf F}^s = {\bf M}^s \otimes {\bf F}^c(T,H,W), \; {\bf M}^s = f^s({\bf F}^c)
\end{equation}
where $f^c()$ and $f^s()$ denote the channel-temporal and spatio-temporal attention functions, respectively.
The arguments of ${\bf F}$ specify the dimensions of element-wise multiplication.
Next we provide the details of the two attention sub-modules.

\subsection{Channel-temporal Attention}
\label{sec:channel_att}
The channel-temporal attention focuses on the `what-when' facets of video attention. Specifically it measures the importance of a particular  object-motion pattern evolving temporally across a video sequence in a specific way. 
For computational efficiency, we squeeze the spatial dimensions ($H\times W$) of each frame-level 3D feature map 
to yield a compact channel descriptor ${\bf d}_{\text{chnl}} \in \mathbb{R}^{T\times C}$
as in \cite{zeiler2014visualizing,hu2018squeeze}.
While average-pooling is a common choice for global spatial information aggregation,
we additionally include max-pooling which would be less likely to 
miss small and/or occluded objects.
Using both pooling operations is also found to be more effective in static image attention modeling \cite{woo2018cbam}.
We denote the two channel descriptors as ${\bf d}_{\text{avg-c}}$
and ${\bf d}_{\text{max-c}} \in \mathbb{R}^{C\times 1 \times 1}$ (indicated by the purple boxes in the top of Fig.~\ref{fig:W3_detail}).

To mine the inter-channel relationships for a given frame,
we then forward ${\bf d}_{\text{avg-c}}$
and ${\bf d}_{\text{max-c}}$ into a shared MLP network $\theta_{\text{c-frm}}$ with one hidden layer
to produce two channel frame attention descriptors, respectively.
We use a bottleneck design with a reduction ratio $r$
which shrinks the hidden activation to the size of $\frac{C}{r}\times 1 \times 1$,
and combine the two frame-level channel attention descriptors by element-wise summation
into a single one ${\bf M}^{\text{c-frm}}$.
We summarize the above {\em frame-level channel-temporal attention} process as
\begin{equation}
    {\bf M}^{\text{c-frm}} = 
    \sigma\Big(f_{\theta_{\text{c-frm}}}({\bf d}_{\text{avg-c}}) \oplus f_{\theta_{\text{c-frm}}}({\bf d}_{\text{max-c}})\Big)
    \in \mathbb{R}^{C\times 1 \times 1},
\end{equation}
where $f_{\theta_{\text{c-frm}}}()$ outputs channel frame attention
and $\sigma()$ is the sigmoid function.

In fine-grained action recognition, temporal dynamics of semantic objects are often the distinguishing factor between classes that involve human interaction with the same object (e.g., opening/closing a book). 
To model the dynamics, a small channel temporal attention network ${\theta_{\text{c-vid}}}$ is introduced, composed of 
a CNN network with two layers of 1D convolutions, to reason about the 
temporally evolving characteristics of each channel dimension (Fig.~\ref{fig:W3_detail} top-right).
This results in our channel-temporal attention mask ${\bf M}^c$, computed as:
\begin{equation}
    {\bf M}^{\text{c}} = 
    \sigma\Big(f_{\theta_{\text{c-vid}}}(\{{\bf M}^{\text{c-frm}}_i\}_{i=1}^T)\Big).
\end{equation}
Concretely, this models the per-channel temporal relationships of successive frames  in a local window specified 
by the kernel size $K_\text{c-vid}$, and composed by two layers %
(we set $K_\text{c-vid}=3$ in our experiments, producing a composed temporal attention span of 6 frames with two 1D CNN layers).
In summary, the parameters of our channel attention model are $\{\theta_{\text{c-frm}}, \theta_{\text{c-vid}}\}$.

\subsection{Spatio-temporal Attention}
\label{sec:spatial_att}
In contrast to the channel-temporal attention that attends to discriminative object local movement patterns evolving temporally in certain ways,  this sub-module attempts to localize them over time.
Similarly, we apply average-pooling and max-pooling along the channel axis to obtain two compact 2D spatial feature maps
for each video frame, denoted as ${\bf d}_{\text{avg-s}}$
and ${\bf d}_{\text{max-s}} \in \mathbb{R}^{1\times H\times W}$.
We then concatenate the two maps and deploy a spatial attention network $\theta_{\text{s-frm}}$ with one 2D convolutional layer 
for each individual frame to output the frame-level spatial attention ${\bf M}^{\text{s-frm}}$.
The kernel size is set to $7\times 7$  (see Fig.~\ref{fig:W3_detail} bottom-left).
To incorporate the temporal dynamics to model how spatial attention evolves over time,
we further perform temporal reasoning on $\{ {\bf M}_i^{\text{s-frm}}\}_{i=1}^T \in \mathbb{R}^{T \times H \times W}$ using a lightweight sub-network $\theta_{\text{s-vid}}$ composed of two 3D convolutional layers.
We adopt the common kernel size of $3\times 3\times 3$~(Fig.~\ref{fig:W3_detail} bottom-right).
We summarize the {\em frame-level and video-level spatial attention} learning as:
\begin{equation}
     {\bf M}^{\text{s-frm}} = 
    \sigma\Big(f_{\theta_{\text{s-frm}}}([{\bf d}_{\text{avg-s}},{\bf d}_{\text{max-s}}])\Big)
    \in \mathbb{R}^{1\times H \times W},
\end{equation}
\begin{equation}
    {\bf M}^{\text{s}} = 
    \sigma\Big(f_{\theta_{\text{s-vid}}}(\{{\bf M}^{\text{s-frm}}_i\}_{i=1}^T)\Big) \in \mathbb{R}^{T \times H \times W}
\end{equation}
The parameters of spatio-temporal attention hence include $\{\theta_{\text{s-frm}}, \theta_{\text{s-vid}}\}$.

\subsection{Model Architecture}
\begin{figure*}[t]
    \centering
    \begin{overpic}[width=0.95\textwidth]{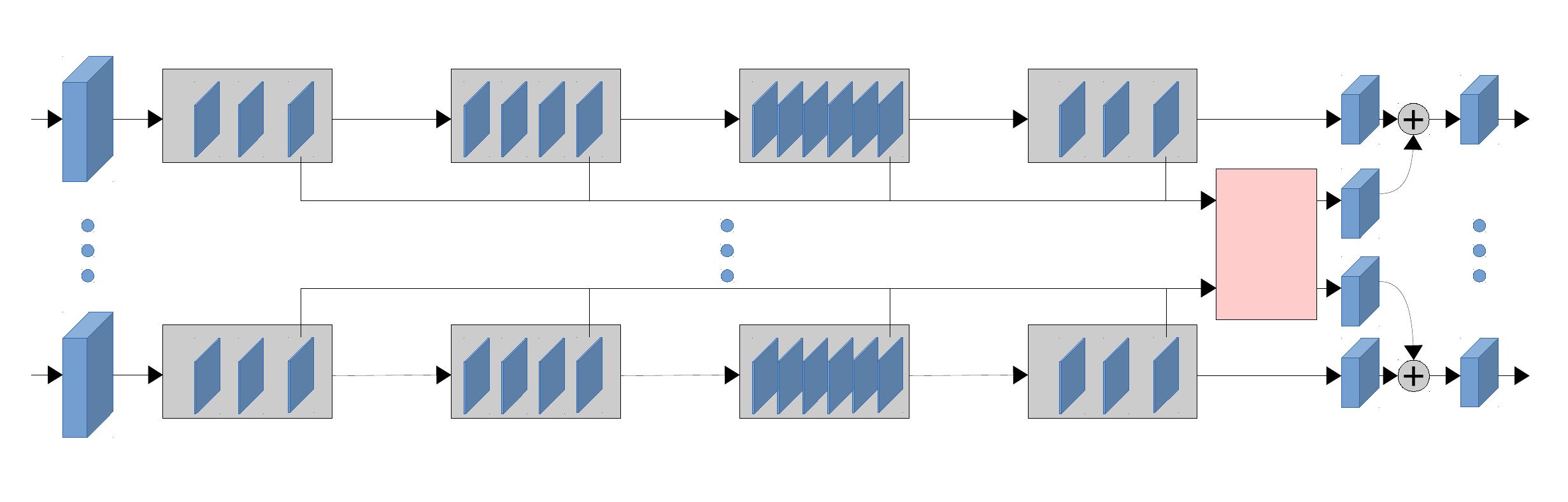}
    
    \put (3,35) {\scriptsize\rotatebox{90}{T frames}}    
    
    \put (10.0,  7.5) {\tiny{Input}}
    \put (42.0,  7.5) {\tiny{Stage I}}
    \put (102.0, 7.5) {\tiny{Stage II}}
    \put (162.0, 7.5) {\tiny{Stage III}}
    \put (222.0, 7.5) {\tiny{Stage IV}}
    
    \put (255.5, 50.5) {\parbox{20pt}{\centering \tiny{$1\times 1$ conv}}}
    
    \put (290.0,  7.5) {\parbox{40pt}{\centering \tiny{Output features}}}
    
    \end{overpic}
    \caption{W3-attention enhanced ResNet-50 architecture
    with the proposed attention guided feature refinement. W3-attention maps are gathered from all
    the ResNet stages, concatenated across the channel dimension, and fed to a 
    $1\times 1$ convolution with ReLU non-linearity. The output is then added to
    the final feature maps.}
    \label{fig:r50}
\end{figure*}
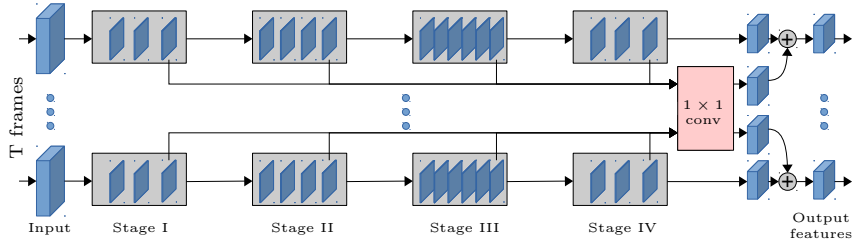

Our \shortname{} video attention module can be easily integrated into any existing CNN architecture. Specifically, it takes as input a 4D feature tensor and outputs an improved same-shape feature tensor with channel-spatio-temporal video attention. In this paper, we focus on the   ResNet-50 based TSM \cite{lin2019tsm} as the main instantiation for integration with \shortname{}. 
Other action models such as I3D \cite{carreira2017quo}
and R2+1D \cite{qiu2017learning,tran2018closer} can be easily integrated without
architectural changes (see Supplementary for more details).
With ResNet-50 as an example,  following the multi-block stage-wise design,
we apply our attention module at each residual block of the backbone, 
i.e., performing the attention learning on every intermediate feature tensor of each stage.
A diagram of W3-attention enhanced ResNet-50 is depicted in Fig.~\ref{fig:r50},
with attention maps for every residual
including a further attention refinement module to be explained in Section~\ref{sec:model}.

\subsection{Model Training}
\label{sec:model}
Learning discriminative video attention would be challenging if trained with standard gradient back propagation through multiple blocks from the top end.
This is because each layer of the action model now has an attention module with temporal reasoning. For those modules, the loss supervision is indirect and
gradually becomes weaker/vanishing when it reaches the bottom levels.
We overcome this issue by exploiting two remedies:
(1) {\em attention guided feature refinement} on architecture design
and 
(2) {\em mature feature-guided regularization} on training strategy. %

\keypoint{Attention guided feature refinement.}
In addition to the standard gradient pathway across the backbone network layers,
we further create another pathway for the attention modules only. Concretely,  we sequentially aggregate all the stage-wise attention masks ${\bf M}_i^{\text{s},j}$ at the frame level, where 
$i$ and $j$ index the frame image and network stage, respectively.
Suppose there are $N$ network stages (e.g., 4 stages in ResNet-50),
we obtain a multi-level attention tensor by adaptive average pooling (AAP) and channel-wise concatenation (Fig.~\ref{fig:r50}):
\begin{equation}
    {\bf M}_i^\text{ms} = [AAP({\bf M}_i^{\text{s},1}), AAP({\bf M}_i^{\text{s},2}), \cdots,
AAP({\bf M}_i^{\text{s},N}) ] \in \mathbb{R}^{N\times H_l \times W_l}
\end{equation}
where $H_l$ and $W_l$ refer to the spatial size of the last stage's feature map ${\bf x}_i \in \mathbb{R}^{C_l \times H_l \times W_l}$.
AAP is for aligning the spatial size of attention masks
from different stages.
Taking ${\bf M}_i^\text{ms}$ as input, we then deploy a tiny CNN network $\theta_\text{ref}$ (composed of one conv layer with $C_l$ $1\times1$ sized kernels for channel size alignment) to produce a feature refining map,
which is further element-wise added to ${\bf x}_i$. %
Formally, it is written as: 
\begin{equation}
  {\bf y}_i =  {\bf x}_i + f_{\theta_\text{ref}}({\bf M}_i^\text{ms}),
  \label{eq:att_ds}
\end{equation}
where ${\bf y}_i$ is the refined feature map of frame $i$.
This process repeats for all the frames of a video sample. 

\textbf{\em Discussion. }
The newly introduced pathway provides
dedicated, joint learning of video attention from multiple stages of the action model backbone and  a shortcut for the gradient flow.
This is because its output is used {\em directly} to aggregate with the final stage feature map, enabling the supervision flowing from the loss down to every single attention module via the shortcuts.
This is essentially a form of  {\em deep supervision} \cite{chang2018multi,lee2015deeply}.

\keypoint{Mature feature-guided regularization.}
Apart from attention deep supervision, we introduce
mature feature guided regularization to further improve the model training.
This follows a two-stage training process.
In the {\em first} stage, we train a video action recognition model with the proposed attention module and attention guided feature refinement (Eq. \eqref{eq:att_ds})
until convergence, and treat it as a teacher model $P$.
In the {\em second} stage, we train the target/student model $Q$ with identical architecture by mimicking the feature maps
of $P$ at the frame level.
Formally, given a frame image $i$ we introduce a feature mimicking regression loss
in the training of $Q$ w.r.t. $P $ as:
\begin{equation}
    \mathcal{L}_{fm} = \| {\bf y}_i^Q - {\bf y}_i^P \|_2
    \label{eq:KD_loss}
\end{equation}
where ${\bf y}_i^Q$ and ${\bf y}_i^P$ are the feature maps
obtained using Eq. \eqref{eq:att_ds} by the target ($Q$) and teacher ($P$) models respectively, with the former serving as the
mature feature to regularize the student's learning process via anchoring to  
a better local optimum than that of the teacher. For the training objective function, we use the summation of cross-entropy classification loss and attention-guided feature refinement loss (Eq. \eqref{eq:att_ds}) in the first stage. The feature mimicking regularization (Eq. \eqref{eq:KD_loss}) further adds up in the second stage. During both training and testing, video-level prediction is obtained by averaging the frame-level predictions.

\textbf{\em Discussion. }
Our regularization is similar to the notion of knowledge distillation (KD) \cite{furlanello2018born,hinton2015distilling,lan2018knowledge,romero2014fitnets,zhang2018deep} but has key differences:
(1) Unlike the conventional KD methods aiming for 
model compression \cite{hinton2015distilling,romero2014fitnets}, we use the same architecture for 
both teacher and student networks.
(2) Compared to \cite{romero2014fitnets}, which also distills feature map knowledge, we only limit to the last attended feature maps rather than multiple ones, and without the need of extra parameters for aligning the feature shape between student and target.
(3) Although \cite{lan2018knowledge,zhang2018deep}
also use the same network architecture for teacher and student, they differently adopt an online distillation strategy which has higher memory usage than our offline counterpart. The representation for distillation used is class prediction distribution which also differs from the feature maps utilized in our model.
(4) Born-Again networks \cite{furlanello2018born}
are more similar to ours since they also perform offline distillation using a single model architecture.
However, as in \cite{lan2018knowledge,zhang2018deep} they also use the class prediction distribution in distillation.
Moreover, only simpler image learning tasks are tested in \cite{furlanello2018born}, whilst we verify our training strategy in a more complex video analysis task.

\section{Experiments}
\label{sec:experiments}

\subsection{Comparisons with Existing State-of-the-Art Methods}

\keypoint{\em Datasets}
We utilized four popular fine-grained action recognition
benchmarks:
{\em Something-Something V1} \cite{goyal2017something}, 
contains $108,499$ videos
from 174 fine-grained action classes 
about hand-object interactions. Some of these classes 
are visually subtle and hence challenging to differentiate, 
such as ``{\em Pretending to turn something upside down}''. 
{\em Something-Something V2} \cite{mahdisoltani2018fine} presents an extended version of V1,
including $220,847$ higher-resolution videos 
with less noisy labels.
{\em EgoGesture}~\cite{zhang2018egogesture} is a 
large scale multi-modal dataset with $21,161$ trimmed videos, 
depicting 83 dynamic gesture classes recorded 
by a head-mounted RGB+D camera in egocentric point of view.
It is challenging due to strong motion blur, heavy
illumination variations, and background clutter
from both indoor and outdoor scenes.
We followed the classification benchmark setting.
{\em EPIC-Kitchens}~\cite{damen2018scaling} is a non-scripted
first-person-view dataset
recorded in 32 real kitchen
environments over multiple days and cities.
It has 55 hours of video, $39,594$ action
segments, $454,255$ object bounding boxes, 
$125$ verb and $331$ noun classes.
We evaluated the classification task 
on verb, noun, and action (verb+noun)
on the standard test set.

\keypoint{\em Training and testing }
We followed the common practice as \cite{lin2019tsm,wang2018non}.
Specifically, the model was trained from 
ImageNet weights
for all the datasets.
For testing, multiple clips are sampled per video and used the full resolution images with shorter side 256.  For efficient inference, we used 1 clip per video and the center crop sized at 224$\times$224.
All the competitors used the same setting for fair comparison.
We reported Top-1/5 accuracy rates for performance evaluation.

\begin{table}[h]
\centering
	\begin{tabular}{c|c|c|c|c|c}
		\toprule %
		{\bf Model } & {\bf Backbone } & {\bf \#Frame } & {\bf GFLOPs} %
		& {\bf Top-1 } & {\bf Top-5 } \\
		\midrule
		TSN \cite{wang2016temporal} & BNI & 8 & 16 & 19.5 & -\\
		TSN* \cite{wang2016temporal} & R50 & 8 & 33 & 19.7 & 46.6\\
		TRN-Multiscale \cite{zhou2018temporal} & BNI & 8 & 16 & 34.4 & -\\
		TRN-Multiscale* \cite{zhou2018temporal} & R50 & 8 & 33 & 38.9 & 68.1\\
		TRN$_\text{RGB+Flow}$ \cite{zhou2018temporal} & BNI & 8+8 & - & 42.0 & -\\
		\midrule
		ECO \cite{zolfaghari2018eco} & BNI+3D R18 & 8 & 32 & 39.6 & -\\
		ECO \cite{zolfaghari2018eco} & BNI+3D R18 & 16 & 64 & 41.4 & -\\
		ECO$_{en}Light$ \cite{zolfaghari2018eco} & BNI+3D R18 & 92 & 267 & 46.4 & -\\
		ECO$_{en}Light_\text{RGB+Flow}$ \cite{zolfaghari2018eco} & BNI+3D R18 & 92+92 & - & 49.5 & -\\
		\midrule
		I3D$^\dagger$ \cite{carreira2017quo} & 3D R50 & 32$\times$2 clip & 153$\times$2 & 41.6 & 72.2 \\
		I3D$^\dagger$+NL \cite{wang2018non} & 3D R50 & 32$\times$2 clip & 168$\times$2 & 44.4 & 76.0 \\
		I3D+NL+GCN \cite{wang2018videos} & 3D R50+GCN & 32$\times$2 clip & 303$\times$2 & 46.1 & 76.8 \\
		SlowFast \cite{feichtenhofer2019slowfast} & (2D+1)R50 & 32 & 65$\times$2 & 47.5 & 76.0 \\
		\midrule
		TSM \cite{lin2019tsm} & R50 & 8 & 33 & 45.6 & 74.2 \\
		TSM \cite{lin2019tsm} & R50 & 16 & 65 & 47.2 & 77.1 \\
		TSM$_{en}$ \cite{lin2019tsm} & R50 & 8+16 & 98 & 49.7 & 78.5\\
		TSM+{\bf \shortname{}} (Ours)
		& R50 & 8 & 33.5 & 49.0 & 77.3 \\
		TSM+{\bf \shortname{}} (Ours)
		& R50 & 16 & 67.1 & \bf 52.6 & \bf 81.3\\
		\bottomrule
	\end{tabular}
	\vspace{.1cm}
	\caption{
	{\bf Comparison with state-of-the-art on Something-Something-V1 \cite{goyal2017something}}.
	Setting: using the center crop and 1 clip per video in test unless otherwise specified.
	BNI=BNInception;
	R18/50=ResNet-18/50;
	$^*$: Results from \cite{lin2019tsm}.
	$^\dagger$: Results from \cite{wang2018videos}.
	}	
	\label{table:SSV1}
\end{table}

\begin{table*}[h]
	\centering
	\begin{tabular}{c|c|c|c|c}
		\toprule
		{\bf Model } & {\bf Backbone }
		& {\bf \# Frames } & {\bf Top-1 } & {\bf Top-5 } \\
		\midrule
		TRN-Multiscale \cite{zhou2018temporal} & BNI & $8\times 2$  & 48.4 & 77.6\\
 		TRN-Multiscale \cite{zhou2018temporal} & R50 & 8 & 38.9 & 68.1\\
		TRN$_\text{RGB+Flow}$ \cite{zhou2018temporal} & BNI & $(8+8)\times 2$ &55.5 & 83.1\\
		\midrule
		TSM \cite{lin2019tsm} & R50 & FR: $8\times 2$ & 59.1 & 85.6\\
		TSM \cite{lin2019tsm} & R50 & FR: $16\times 2$ & 63.1 & 88.1\\
		\midrule
		TSM+{\bf \shortname{}}
		& R50 & CC: $16\times 2$ &\bf 65.7 &\bf 90.2 \\
		TSM+{\bf \shortname{}}
		& R50 & FR: $16\times 2$ &\bf 66.5 &\bf 90.4 \\		
		\bottomrule
	\end{tabular}
	\vspace{.1cm}
	\caption{
	{\bf Comparison with state-of-the-art on Something-Something-V2 \cite{mahdisoltani2018fine}}. 
	FR=Full Resolution testing;
	CC=Center Crop testing.
	In testing, two clips per video were used.
	}
	\label{table:SSV2}
\end{table*}

\keypoint{Results on Something-Something V1} 
We compared our \shortname{} method with the state-of-the-art competitors in Table~\ref{table:SSV1}.
It is evident that our \shortname{} with TSM \cite{lin2019tsm} yields the best results among all the competitors, which validates the overall performance superiority of our attention model.
We summarize detailed comparisons as below:
{\bf (1)} {\em 2D models} ($1^\text{st}$ block):
Without temporal inference, 2D models such as TSN \cite{wang2016temporal} perform the worst, as expected. Whilst the performance can be
improved clearly using independent temporal modeling after feature extraction with TRN \cite{zhou2018temporal}, it remains much lower than the recent 3D models.
{\bf (2)} {\em 2D+3D models} ($2^\text{nd}$ block):
As shown for ECO, the introduction of 3D spatio-temporal feature extraction notably boosts the performance w.r.t. TSN and TRN.
However, these methods still suffer from high computational cost for a model with competitive performance.
{\bf (3)} {\em 3D models} ($3^\text{rd}$ block):
The I3D model \cite{carreira2017quo} has been considered widely as a strong baseline and further improvements have been added in \cite{feichtenhofer2019slowfast,wang2018non,wang2018videos} including self-attention based non-local network. 
A clear weakness of these methods
is their huge computational cost, making deployment on resource-constrained devices impossible.
{\bf (4)} {\em Time-shift models} ($4^\text{th}$ block): As the previous state-of-the-art model, TSM \cite{lin2019tsm} yields remarkable accuracy with the computational cost as low as 2D models. 
Importantly, our \shortname{} attention on top of TSM 
further boosts the performance by a significant margin.
For instance, it
achieves a Top-1 gain of 3.6\%/5.4\% when using 8/16 frames per video in test, with only a small extra cost of 0.5G/2.1G FLOPs.

\keypoint{Results on Something-Something V2}
The results are shown  in Table \ref{table:SSV2}.
Following \cite{lin2019tsm},
we used two clips per video each with 16 frames
in testing.
Overall, we have similar observation as on V1.
For instance, TRN \cite{zhou2018temporal} is clearly inferior to our baseline TSM \cite{lin2019tsm}, and our method further significantly improves the Top-1 accuracy by 3.4\% when using 16$\times$2 full resolution frames.

\begin{table*}[h]
	\resizebox{0.98\linewidth}{!}{%
	\begin{tabular}{c|c|c|c|c|c|c}
		\toprule
		{\bf Model}  & {\bf Backbone } & {\bf \#Frames} & {\bf Val Top-1} & {\bf Val Top-5} & {\bf Test Top-1} & {\bf Test Top-5} \\
		\midrule
        VGG16+LSTM \cite{graves2013generating} & VGG16 & 16 & - & - &
        74.7 & -\\
        \midrule
        C3D \cite{tran2015learning} & C3D  & 8 & - & - & 81.7 & -\\
        C3D \cite{tran2015learning} & C3D & 16 & - & - & 86.4 & -\\
        C3D+LSTM+RSTM \cite{cao2017egocentric} & C3D & 16 & - & - & 89.3 & -\\
        \midrule
        I3D \cite{carreira2017quo} & I3D & 32 & - & - & 90.3 & -\\
        \midrule
		TSM \cite{lin2019tsm} & R50 & 8 & 79.7 & 96.9 & 80.5 & 97.8\\
		TSM+{\bf \shortname{}}
		& R50 & 8 & \bf{93.9} & \bf{98.7} & \bf{94.3} & \bf{99.2}
		\\
		\bottomrule
	\end{tabular}
	}
	\vspace{.1cm}
	\caption{
	{\bf Comparison with state-of-the-art on EgoGesture \cite{zhang2018egogesture}}. 
	Setting: 1 crop per video in test, using RGB frames only unless specified otherwise.
	}
	\label{table:ego-gesture}
\end{table*}

\keypoint{Results on EgoGesture} 
Table~\ref{table:ego-gesture} shows comparative results on this ego-centric dataset.
For fast inference, we used only 8 RGB frames per video
for TSM and our model.
It is obvious that our \shortname{} with TSM outperforms all the competitors, often with a large margin.

\begin{table*}[h]%
	\centering
	\setlength{\tabcolsep}{0.1cm} 
	\resizebox{\columnwidth}{!}{%
	\begin{tabular}{c|cc|cc|cc|cc|cc|cc}
		\toprule
		 & \multicolumn{4}{c|}{ {\bf Verb} } & \multicolumn{4}{c|}{ {\bf Noun} }  
		& \multicolumn{4}{c}{ {\bf Action} }  \\
		& \multicolumn{2}{c|}{Top-1} & \multicolumn{2}{c|}{Top-5}  
		& \multicolumn{2}{c|}{Top-1} & \multicolumn{2}{c|}{Top-5}
		& \multicolumn{2}{c|}{Top-1} & \multicolumn{2}{c}{Top-5} \\
		{\bf Model} & S1 & S2 & S1 & S2 & S1 & S2 & S1 & S2 & S1 & S2 & S1 & S2  \\
		\midrule
		TSN$^*$ \cite{wang2016temporal} 
		& 49.7 & 36.7 & 87.2 & 73.6 & 39.9 & 23.1 & 65.9 & 44.7 & 24.0 & 12.8 & 46.1 & 26.1 \\
		TRN$^*$ \cite{zhou2018temporal} 
		& 58.8 & 47.3 & 86.6 & 76.9 & 37.3 & 23.7 & 63.0 & 46.0 & 26.6 & 15.7 & 46.1 & 30.0 \\
		TRN-Multiscale$^*$ \cite{zhou2018temporal} 
		& 60.2 & 46.9 & 87.2 & 75.2 & 38.4 & 24.4 & 64.7 & 46.7 & 28.2 & 16.3 & 47.9 & 29.7 \\
		Action Banks~\cite{wu2019long} - full res & 60.0 & 50.9 & 88.4 & 77.6 &\bf 45.0 &\bf 31.5 &\bf 71.8 &\bf 57.8 & 32.7 & \bf 21.2 &\bf 55.3 &\bf 39.4 \\
		\midrule
		TSM$^*$ \cite{lin2019tsm} 
		& 57.9 & 43.5 & 87.1 & 73.9 & 40.8 & 23.3 & 66.1 & 46.0 & 28.2 & 15.0 & 49.1 & 28.1 \\

		TSM+{\bf \shortname{}} %
		&64.4 & 50.2 &\bf 88.8 & 78.0 & 44.2 & 26.6 & 68.1 & 49.5 & 33.5 & 17.8 & 53.9 & 32.6
		\\

		TSM+{\bf \shortname{}} - full res %
		&\bf  64.7 &\bf  51.4 &\bf 88.8 &\bf 78.5 & 44.7 & 27.0 & 69.0 & 50.3 &\bf 34.2 & 18.7 & 54.6 & 33.7
		\\
		\bottomrule
	\end{tabular}
	}
	\vspace{.1cm}
	\caption{
	{\bf Comparison with state-of-the-art on EPIC-Kitchens \cite{damen2018scaling}}. 
	Setting: 8 frames per video and 10 crops in test and only RGB frames.
	S1: Seen Kitchens;
	S2: Unseen Kitchens.
	`$^*$': Results from \cite{price2019evaluation}.
	}
	\label{table:epick}
\end{table*}

\keypoint{Results on EPIC-Kitchens} 
We compared our method with a number of 
state-of-the-art action models
in Table~\ref{table:epick}.
In this experiment, we adopted the test setup of \cite{price2019evaluation}:
using two clips per video and ten crops.
On this realistic and challenging dataset, we observed consistent performance gain
obtained by adding our \shortname{} attention model to the baseline TSM
across verb, noun, and action classification.
This leads to the best accuracy rates among all the 
strong competitors evaluated in the same setting.
For example, \shortname{} improves the action top-1 accuracy by 5.3\%/2.8\% on seen/unseen kitchen test sets.
We also report a clear margin over the current state-of-the-art model,
Action Banks \cite{wu2019long} on verb classification. 
Note that Action Banks uses more temporal data for every action and noun prediction, and an extra object detector. 
This gives it an unfair advantage over our model, and  explains its better performance on noun classification and subsequent action classification.  
The results validate the importance of spatio-temporal attention learning for action recognition in unconstrained egocentric videos,
and the effectiveness of our \shortname{} attention formulation.

\subsection{Ablation Study}

\keypoint{\em Setting}
We conducted an in-depth ablation study of our \shortname{} attention module
on Something-Something V1 \cite{goyal2017something}, 
We used ResNet-50 based TSM \cite{lin2019tsm} as the baseline and adopted the same setting.
We fine-tuned the model from ImageNet pre-trained weights with 16 RGB frames per video.
In testing, we used 1 clip per video
and the center crop of $224\times224$.
We adopted the Top-1 accuracy as the performance evaluation metric.

\begin{table}[h]
    \begin{minipage}{.55\linewidth}
    \scriptsize
      \centering
      \resizebox{\columnwidth}{!}{%
        \begin{tabular}{c|c|c|c|c|c}
		\toprule
		 & Full & w/o MFR & w/o AFR & w/o SA & w/o TA \\
		\midrule
		Top-1 & \bf  52.6 & 50.3 & 50.1 & 49.8 & 47.2\\
		\bottomrule
	\end{tabular}
	}
	\vspace{.1cm}
	\caption{
	{\bf \scriptsize Model component analysis.}
	\scriptsize{
	MFR=Mature Feature guided regularization;
	AFR=Attention-guided Feature Refinement;
	SA=Spatial Attention; 
	TA=Temporal Attention.
	}
	}
	\label{tab:component}
    \end{minipage}%
    \hfill
    \begin{minipage}{.4\linewidth}
      \centering
      \resizebox{\columnwidth}{!}{%
        \begin{tabular}{c|c|c|c|c}
		\toprule
		& TSM  & +CBAM \cite{woo2018cbam} & +NL \cite{wang2018non}
		& +{\bf \shortname{}}  \\
		\midrule
		Top-1 & 47.2 & 49.1 & 49.8 &\bf 52.6 \\
        GFLOPs & 65.0 & 66.5 & 115.0 & 67.1 \\
		\bottomrule
	\end{tabular}
	}
	\vspace{.1cm}
	\caption{
	{\bf \scriptsize Comparing attention models}. %
	\scriptsize{
	NL=Non Local;
	\newline
	\newline
	}
	}
	\label{tab:att}
    \end{minipage} 
\end{table}

\keypoint{Model component analysis}
In Table \ref{tab:component}, we examined the effect of three main components in our \shortname{} attention
by removing them one at a time
due to the dependence nature in design.
We observed that:
(i) In model optimization,
our mature feature guided regularization brings the majority performance gain,
whilst attention guided feature refinement also helps improve the results.
(ii) Among three attention facets, if we keep the channel (`what') facet and remove one of the other two, we can see that 
temporal attention turns out to be more important than
spatial attention. This reflects the fundamental difference
between video and image analysis tasks.

\keypoint{Comparing attention models}
We compared our \shortname{} attention model with
two strong competitors: 
(1) CBAM \cite{woo2018cbam} which is the state-of-the-art image attention;
(2) Non-Local operator \cite{wang2018non} which is the best  video attention module thus far in the literature.
The results in Table \ref{tab:att} show that
(i) Our \shortname{} attention
yields the most significant accuracy boost over the base action model TSM \cite{lin2019tsm},
validating the overall performance advantages of our attention operation.
(ii) When combined with TSM and end-to-end trained, surprisingly CBAM produces a very strong performance
on par with Non-Local attention, indicating that a strong video action method can be composited by
simply applying image attention to top-performing action models. 
However, there is still a clear gap against the proposed \shortname{}
with a more principled way of learning spatio-temporal video attention. 
(iii) \shortname{} achieves this by being much less compute-intensive than the Non-local alternative, adding virtually no extra computational cost on top of TSM. 
The above analysis suggests that CBAM and our mature feature guided regularization (MFR) both
are very effective. We thus expect their combination to be a stronger competitor.
The results in Table \ref{tab:CBAM_KD} validate this  --
an extra +1.4\% boost in Top-1 accuracy over CBAM alone when our new training strategy is applied.
Interestingly, we note that the gain by MFR is more significant (2.4\% increase)
when used with the proposed attention mechanism.
A plausible reason is that richer spatio-temporal information is captured 
by our attention model in the first training stage, 
which further benefits the subsequent regularization process.

\begin{table}[t]
      \centering
      \resizebox{\columnwidth}{!}{%
        \begin{tabular}{m{15mm}|c||c|c|m{15mm}||c|c|m{15mm}}
		\toprule
		 & TSM  & + CBAM \cite{woo2018cbam} & + CBAM w/ MFR &\bf\em Gain
		& + {\bf \shortname{}} w/o MFR  & + {\bf \shortname{}} &\bf\em  Gain \\
		\midrule
		Top-1 & 47.2 & 49.1 & 50.5 & +1.4 & 50.2 & 52.6 &\bf +2.4 \\
		\bottomrule
	\end{tabular}
	}
	\vspace{.1cm}
	\caption{
	{\bf Impact of our Mature Feature guided Regularization (MFR).}
	Refer to text for details.
	}
	\label{tab:CBAM_KD}
\end{table}

\begin{figure*}[th!]
   \centering
    \begin{tabular}{c ccccc}
    \scriptsize\rotatebox{90}{~~~~RGB}&
    \includegraphics[width=0.167\linewidth]{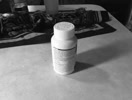}&
    \includegraphics[width=0.167\linewidth]{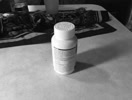}&
    \includegraphics[width=0.167\linewidth]{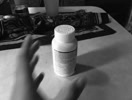}&
    \includegraphics[width=0.167\linewidth]{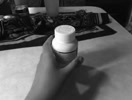}&
    \includegraphics[width=0.167\linewidth]{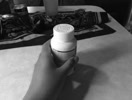}\\    
    
    \scriptsize\rotatebox{90}{~~~~CBAM}&
    \includegraphics[width=0.167\linewidth]{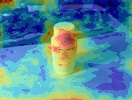}&
    \includegraphics[width=0.167\linewidth]{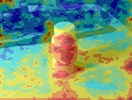}&
    \includegraphics[width=0.167\linewidth]{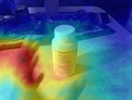}&
    \includegraphics[width=0.167\linewidth]{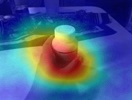}&
    \includegraphics[width=0.167\linewidth]{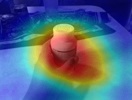}\\
    
    \scriptsize\rotatebox{90}{~~~~~~W3}&
    \includegraphics[width=0.167\linewidth]{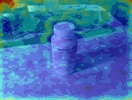}&
    \includegraphics[width=0.167\linewidth]{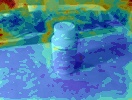}&
    \includegraphics[width=0.167\linewidth]{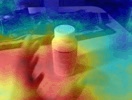}&
    \includegraphics[width=0.167\linewidth]{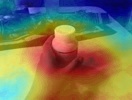}&
    \includegraphics[width=0.167\linewidth]{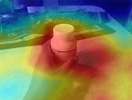}\\    
    \\
    \scriptsize\rotatebox{90}{~~~~RGB}&
    \includegraphics[width=0.167\linewidth]{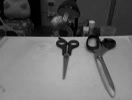}&
    \includegraphics[width=0.167\linewidth]{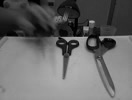}&
    \includegraphics[width=0.167\linewidth]{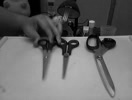}&
    \includegraphics[width=0.167\linewidth]{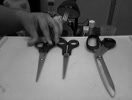}&
    \includegraphics[width=0.167\linewidth]{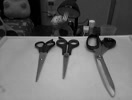}\\    
    
    \scriptsize\rotatebox{90}{~~~~CBAM}&
    \includegraphics[width=0.167\linewidth]{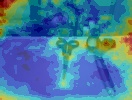}&
    \includegraphics[width=0.167\linewidth]{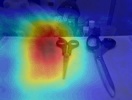}&
    \includegraphics[width=0.167\linewidth]{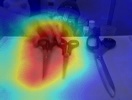}&
    \includegraphics[width=0.167\linewidth]{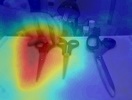}&
    \includegraphics[width=0.167\linewidth]{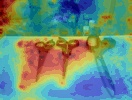}\\
    
    \scriptsize\rotatebox{90}{~~~~~~W3}&
    \includegraphics[width=0.167\linewidth]{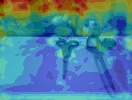}&
    \includegraphics[width=0.167\linewidth]{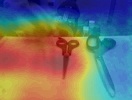}&
    \includegraphics[width=0.167\linewidth]{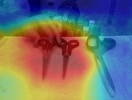}&
    \includegraphics[width=0.167\linewidth]{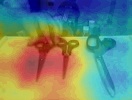}&
    \includegraphics[width=0.167\linewidth]{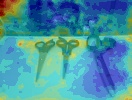}\\     
    \\
    \scriptsize\rotatebox{90}{~~~~~RGB}&
    \includegraphics[width=0.167\linewidth]{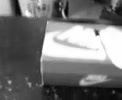}&
    \includegraphics[width=0.167\linewidth]{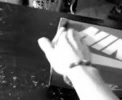}&
    \includegraphics[width=0.167\linewidth]{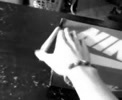}&
    \includegraphics[width=0.167\linewidth]{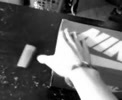}&
    \includegraphics[width=0.167\linewidth]{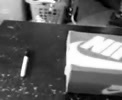}\\    
    
    \scriptsize\rotatebox{90}{~~~~~CBAM}&
    \includegraphics[width=0.167\linewidth]{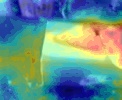}&
    \includegraphics[width=0.167\linewidth]{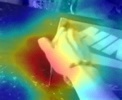}&
    \includegraphics[width=0.167\linewidth]{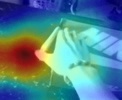}&
    \includegraphics[width=0.167\linewidth]{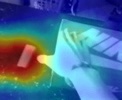}&
    \includegraphics[width=0.167\linewidth]{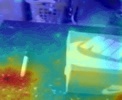}\\
    
    \scriptsize\rotatebox{90}{~~~~~~W3}&
    \includegraphics[width=0.167\linewidth]{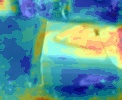}&
    \includegraphics[width=0.167\linewidth]{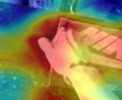}&
    \includegraphics[width=0.167\linewidth]{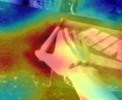}&
    \includegraphics[width=0.167\linewidth]{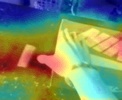}&
    \includegraphics[width=0.167\linewidth]{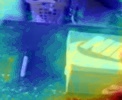}\\
    
    \end{tabular}
    \caption{
    {\bf Comparison of spatio-temporal attention regions.}
    Attention masks by CBAM \cite{woo2018cbam} and our method on Something-Something V1.
    }
    \label{fig:teaser} 
\end{figure*}

\keypoint{Attention visualization and interpretation}
We visualized the attended regions of our \shortname{} in comparison to CBAM.
The attention masks of three video sequences each with five evenly sampled frames are shown in Fig.~\ref{fig:teaser}.
We observed that,
CBAM tends to attend objects and/or hands whenever 
they appear, regardless if some action is happening or not
(i.e., static spatial attention).
In contrast, our \shortname{} attention is activated only when
a specific action is ongoing (i.e., spatio-temporal attention), and importantly 
all the informative regions are attended including 
hands, active objects, and related scene context.
For instance, in the second example action 
{\em ``Placing something next to something''},
the neighbouring scissor is also attended,
which clearly interprets how our model can make a correct prediction.
A similar phenomenon is observed in the third example 
action {\em ``Pushing something off something''}.

\section{Conclusions}
We have presented a novel lightweight video attention module
for fine-grained action recognition in unconstrained videos.
Used simply as a drop-in building block,
our proposed \shortname{} module significantly improves the
performance of existing action recognition methods
with very small extra overhead.
It yields superior performance over a number of state-of-the-art 
alternatives on a variety of action recognition tasks.

\clearpage
\bibliographystyle{splncs04}
\bibliography{egbib}

\begin{thebibliography}{10}
\providecommand{\url}[1]{\texttt{#1}}
\providecommand{\urlprefix}{URL }
\providecommand{\doi}[1]{https://doi.org/#1}

\bibitem{baccouche2011sequential}
Baccouche, M., Mamalet, F., Wolf, C., Garcia, C., Baskurt, A.: Sequential deep
  learning for human action recognition. In: International Workshop on Human
  Behavior Understanding (2011)

\bibitem{bahdanau2014neural}
Bahdanau, D., Cho, K., Bengio, Y.: Neural machine translation by jointly
  learning to align and translate. In: International Conference on Learning
  Representations (2015)

\bibitem{cao2017egocentric}
Cao, C., Zhang, Y., Wu, Y., Lu, H., Cheng, J.: Egocentric gesture recognition
  using recurrent 3d convolutional neural networks with spatiotemporal
  transformer modules. In: IEEE International Conference on Computer Vision.
  pp. 3763--3771 (2017)

\bibitem{carreira2017quo}
Carreira, J., Zisserman, A.: Quo vadis, action recognition? a new model and the
  kinetics dataset. In: IEEE Conference on Computer Vision and Pattern
  Recognition. pp. 6299--6308 (2017)

\bibitem{chang2018multi}
Chang, X., Hospedales, T.M., Xiang, T.: Multi-level factorisation net for
  person re-identification. In: IEEE Conference on Computer Vision and Pattern
  Recognition. pp. 2109--2118 (2018)

\bibitem{damen2018scaling}
Damen, D., Doughty, H., Maria~Farinella, G., Fidler, S., Furnari, A., Kazakos,
  E., Moltisanti, D., Munro, J., Perrett, T., Price, W., et~al.: Scaling
  egocentric vision: The epic-kitchens dataset. In: European Conference on
  Computer Vision. pp. 720--736 (2018)

\bibitem{devlin-etal-2019-bert}
Devlin, J., Chang, M.W., Lee, K., Toutanova, K.: {BERT}: Pre-training of deep
  bidirectional transformers for language understanding. In: ACL. pp.
  4171--4186 (2019)

\bibitem{donahue2015long}
Donahue, J., Anne~Hendricks, L., Guadarrama, S., Rohrbach, M., Venugopalan, S.,
  Saenko, K., Darrell, T.: Long-term recurrent convolutional networks for
  visual recognition and description. In: IEEE Conference on Computer Vision
  and Pattern Recognition (2015)

\bibitem{du2017recurrent}
Du, W., Wang, Y., Qiao, Y.: Recurrent spatial-temporal attention network for
  action recognition in videos. T-IP  \textbf{27}(3),  1347--1360 (2017)

\bibitem{MoreIsLess19}
Fan, Q., Chen, C.F.R., Kuehne, H., Pistoia, M., Cox, D.: More is less: Learning
  efficient video representations by big-little network and depthwise temporal
  aggregation. In: Advances on Neural Information Processing Systems (2019)

\bibitem{feichtenhofer2019slowfast}
Feichtenhofer, C., Fan, H., Malik, J., He, K.: {SlowFast} networks for video
  recognition. In: IEEE International Conference on Computer Vision (2019)

\bibitem{fu2019dual}
Fu, J., Liu, J., Tian, H., Li, Y., Bao, Y., Fang, Z., Lu, H.: Dual attention
  network for scene segmentation. In: IEEE Conference on Computer Vision and
  Pattern Recognition (2019)

\bibitem{furlanello2018born}
Furlanello, T., Lipton, Z.C., Tschannen, M., Itti, L., Anandkumar, A.: Born
  again neural networks. In: International Conference on Machine Learning
  (2018)

\bibitem{gehring2017convolutional}
Gehring, J., Auli, M., Grangier, D., Yarats, D., Dauphin, Y.N.: Convolutional
  sequence to sequence learning. In: International Conference on Machine
  Learning (2017)

\bibitem{girdhar2019video}
Girdhar, R., Carreira, J., Doersch, C., Zisserman, A.: Video action transformer
  network. In: IEEE Conference on Computer Vision and Pattern Recognition. pp.
  244--253 (2019)

\bibitem{girdhar2017attentional}
Girdhar, R., Ramanan, D.: Attentional pooling for action recognition. In:
  Advances on Neural Information Processing Systems. pp. 34--45 (2017)

\bibitem{goyal2017something}
Goyal, R., Kahou, S.E., Michalski, V., Materzynska, J., Westphal, S., Kim, H.,
  Haenel, V., Fruend, I., Yianilos, P., Mueller-Freitag, M., et~al.: The
  ``something something'' video database for learning and evaluating visual
  common sense. In: IEEE International Conference on Computer Vision (2017)

\bibitem{graves2013generating}
Graves, A.: Generating sequences with recurrent neural networks. arXiv preprint
  arXiv:1308.0850  (2013)

\bibitem{gu2018learning}
Gu, J., Hu, H., Wang, L., Wei, Y., Dai, J.: Learning region features for object
  detection. In: European Conference on Computer Vision (2018)

\bibitem{hinton2015distilling}
Hinton, G., Vinyals, O., Dean, J.: Distilling the knowledge in a neural
  network. arXiv preprint arXiv:1503.02531  (2015)

\bibitem{hu2018squeeze}
Hu, J., Shen, L., Sun, G.: Squeeze-and-excitation networks. In: IEEE
  International Conference on Computer Vision (2018)

\bibitem{huang2019ccnet}
Huang, Z., Wang, X., Huang, L., Huang, C., Wei, Y., Liu, W.: Ccnet: Criss-cross
  attention for semantic segmentation. In: IEEE International Conference on
  Computer Vision. pp. 603--612 (2019)

\bibitem{ji20123d}
Ji, S., Xu, W., Yang, M., Yu, K.: 3d convolutional neural networks for human
  action recognition. IEEE Transactions of Pattern Analysis and Machine
  Intelligence  \textbf{35}(1),  221--231 (2012)

\bibitem{karpathy2014large}
Karpathy, A., Toderici, G., Shetty, S., Leung, T., Sukthankar, R., Fei-Fei, L.:
  Large-scale video classification with convolutional neural networks. In:
  CVPR. pp. 1725--1732 (2014)

\bibitem{lan2018knowledge}
Lan, X., Zhu, X., Gong, S.: Knowledge distillation by on-the-fly native
  ensemble. In: Advances on Neural Information Processing Systems. pp.
  7517--7527 (2018)

\bibitem{lee2015deeply}
Lee, C.Y., Xie, S., Gallagher, P., Zhang, Z., Tu, Z.: Deeply-supervised nets.
  In: Artificial Intelligence and Statistics. pp. 562--570 (2015)

\bibitem{li2018eye}
Li, Y., Liu, M., Rehg, J.M.: In the eye of beholder: Joint learning of gaze and
  actions in first person video. In: European Conference on Computer Vision
  (2018)

\bibitem{li2018videolstm}
Li, Z., Gavrilyuk, K., Gavves, E., Jain, M., Snoek, C.G.: Videolstm convolves,
  attends and flows for action recognition. CVIU  \textbf{166},  41--50 (2018)

\bibitem{lin2019tsm}
Lin, J., Gan, C., Han, S.: Tsm: Temporal shift module for efficient video
  understanding. In: IEEE International Conference on Computer Vision. pp.
  7083--7093 (2019)

\bibitem{luong2015effective}
Luong, M.T., Pham, H., Manning, C.D.: Effective approaches to attention-based
  neural machine translation. arXiv preprint arXiv:1508.04025  (2015)

\bibitem{mahdisoltani2018fine}
Mahdisoltani, F., Berger, G., Gharbieh, W., Fleet, D., Memisevic, R.:
  Fine-grained video classification and captioning. arXiv preprint
  arXiv:1804.09235  \textbf{5}(6) (2018)

\bibitem{Martinez_2019_ICCV}
Martinez, B., Modolo, D., Xiong, Y., Tighe, J.: Action recognition with
  spatial-temporal discriminative filter banks. In: IEEE International
  Conference on Computer Vision (2019)

\bibitem{meng2019interpretable}
Meng, L., Zhao, B., Chang, B., Huang, G., Sun, W., Tung, F., Sigal, L.:
  Interpretable spatio-temporal attention for video action recognition. In:
  IEEE International Conference on Computer Vision Workshop. pp.~0--0 (2019)

\bibitem{park2018bam}
Park, J., Woo, S., Lee, J.Y., Kweon, I.S.: Bam: Bottleneck attention module.
  In: British Machine Vision Conference (2018)

\bibitem{price2019evaluation}
Price, W., Damen, D.: An evaluation of action recognition models on
  epic-kitchens. arXiv preprint arXiv:1908.00867  (2019)

\bibitem{qiu2017learning}
Qiu, Z., Yao, T., Mei, T.: Learning spatio-temporal representation with
  pseudo-{3D} residual networks. In: IEEE International Conference on Computer
  Vision (2017)

\bibitem{Rensink2000humanAttention}
Rensink, R.A.: The dynamic representation of scenes. Visual Cognition
  \textbf{1}(1--3),  17--42 (2000)

\bibitem{romero2014fitnets}
Romero, A., Ballas, N., Kahou, S.E., Chassang, A., Gatta, C., Bengio, Y.:
  Fitnets: Hints for thin deep nets. arXiv preprint arXiv:1412.6550  (2014)

\bibitem{sharma2016action}
Sharma, S., Kiros, R., Salakhutdinov, R.: Action recognition using visual
  attention. In: International Conference on Learning Representations Workshop
  (2016)

\bibitem{si2019attention}
Si, C., Chen, W., Wang, W., Wang, L., Tan, T.: An attention enhanced graph
  convolutional lstm network for skeleton-based action recognition. In: IEEE
  Conference on Computer Vision and Pattern Recognition. pp. 1227--1236 (2019)

\bibitem{simonyan2014two}
Simonyan, K., Zisserman, A.: Two-stream convolutional networks for action
  recognition in videos. In: Advances in neural information processing systems.
  pp. 568--576 (2014)

\bibitem{song2017end}
Song, S., Lan, C., Xing, J., Zeng, W., Liu, J.: An end-to-end spatio-temporal
  attention model for human action recognition from skeleton data. In: AAAI
  (2017)

\bibitem{soomro2012ucf101}
Soomro, K., Zamir, A.R., Shah, M.: Ucf101: A dataset of 101 human actions
  classes from videos in the wild. arXiv preprint arXiv:1212.0402  (2012)

\bibitem{sudhakaran2018attention}
Sudhakaran, S., Lanz, O.: Attention is all we need: Nailing down object-centric
  attention for egocentric activity recognition. In: British Machine Vision
  Conference (2018)

\bibitem{Tao2018nonlocalproblem}
Tao, Y., Sun, Q., Du, Q., Liu, W.: Nonlocal neural networks, nonlocal diffusion
  and nonlocal modeling. In: Advances on Neural Information Processing Systems
  (2018)

\bibitem{torabi2017action}
Torabi, A., Sigal, L.: Action classification and highlighting in videos. arXiv
  preprint arXiv:1708.09522  (2017)

\bibitem{tran2015learning}
Tran, D., Bourdev, L., Fergus, R., Torresani, L., Paluri, M.: Learning
  spatiotemporal features with 3d convolutional networks. In: IEEE
  International Conference on Computer Vision (2015)

\bibitem{Tran2019channel-separated}
Tran, D., Wang, H., Torresani, L., Feiszli, M.: Video classification with
  channel-separated convolutional networks. In: IEEE International Conference
  on Computer Vision (2019)

\bibitem{tran2018closer}
Tran, D., Wang, H., Torresani, L., Ray, J., LeCun, Y., Paluri, M.: A closer
  look at spatiotemporal convolutions for action recognition. In: IEEE
  Conference on Computer Vision and Pattern Recognition (2018)

\bibitem{vaswani2017attention}
Vaswani, A., Shazeer, N., Parmar, N., Uszkoreit, J., Jones, L., Gomez, A.N.,
  Kaiser, {\L}., Polosukhin, I.: Attention is all you need. In: Advances on
  Neural Information Processing Systems (2017)

\bibitem{wang2018action}
Wang, L., Zang, J., Zhang, Q., Niu, Z., Hua, G., Zheng, N.: Action recognition
  by an attention-aware temporal weighted convolutional neural network. Sensors
   \textbf{18}(7), ~1979 (2018)

\bibitem{wang2016temporal}
Wang, L., Xiong, Y., Wang, Z., Qiao, Y., Lin, D., Tang, X., Van~Gool, L.:
  Temporal segment networks: Towards good practices for deep action
  recognition. In: European Conference on Computer Vision (2016)

\bibitem{wang2018non}
Wang, X., Girshick, R., Gupta, A., He, K.: Non-local neural networks. In: IEEE
  Conference on Computer Vision and Pattern Recognition (2018)

\bibitem{wang2018videos}
Wang, X., Gupta, A.: Videos as space-time region graphs. In: European
  Conference on Computer Vision (2018)

\bibitem{wang2016hierarchical}
Wang, Y., Wang, S., Tang, J., O'Hare, N., Chang, Y., Li, B.: Hierarchical
  attention network for action recognition in videos. arXiv preprint
  arXiv:1607.06416  (2016)

\bibitem{woo2018cbam}
Woo, S., Park, J., Lee, J.Y., So~Kweon, I.: Cbam: Convolutional block attention
  module. In: European Conference on Computer Vision. pp. 3--19 (2018)

\bibitem{wu2019long}
Wu, C.Y., Feichtenhofer, C., Fan, H., He, K., Krahenbuhl, P., Girshick, R.:
  Long-term feature banks for detailed video understanding. In: IEEE Conference
  on Computer Vision and Pattern Recognition (2019)

\bibitem{yang2018action}
Yang, Z., Li, Y., Yang, J., Luo, J.: Action recognition with spatio--temporal
  visual attention on skeleton image sequences. T-CSVT  \textbf{29}(8),
  2405--2415 (2018)

\bibitem{yue2015beyond}
Yue-Hei~Ng, J., Hausknecht, M., Vijayanarasimhan, S., Vinyals, O., Monga, R.,
  Toderici, G.: Beyond short snippets: Deep networks for video classification.
  In: CVPR. pp. 4694--4702 (2015)

\bibitem{zeiler2014visualizing}
Zeiler, M.D., Fergus, R.: Visualizing and understanding convolutional networks.
  In: European Conference on Computer Vision. pp. 818--833 (2014)

\bibitem{zhang2018egogesture}
Zhang, Y., Cao, C., Cheng, J., Lu, H.: Egogesture: a new dataset and benchmark
  for egocentric hand gesture recognition. T-Multimedia  \textbf{20}(5),
  1038--1050 (2018)

\bibitem{zhang2018deep}
Zhang, Y., Xiang, T., Hospedales, T.M., Lu, H.: Deep mutual learning. In: IEEE
  Conference on Computer Vision and Pattern Recognition. pp. 4320--4328 (2018)

\bibitem{zhou2018temporal}
Zhou, B., Andonian, A., Oliva, A., Torralba, A.: Temporal relational reasoning
  in videos. In: European Conference on Computer Vision (2018)

\bibitem{zolfaghari2018eco}
Zolfaghari, M., Singh, K., Brox, T.: Eco: Efficient convolutional network for
  online video understanding. In: European Conference on Computer Vision. pp.
  695--712 (2018)

\end{thebibliography}

\newpage
\section{Supplementary: Experiments with other backbones}

\begin{table}[h]
\centering
	\begin{tabular}{c|c|c|c|c}
		\toprule %
		{\bf Model } & {\bf Backbone } & {\bf \#Frame }
		& {\bf Top-1 } & {\bf Top-5 } \\
		\midrule
		\midrule
		I3D  \cite{carreira2017quo}                                 & 3D R18 & 32 & 34.9 & 62.6 \\
		I3D+{\bf \shortname{}} w/o MFR (Ours)  & 3D R18 & 32 & 35.3 & 63.5 \\
        \midrule
		I3D  \cite{carreira2017quo}                                 & 3D R18 & 64 & 36.4  & 63.7 \\
		I3D+{\bf \shortname{}} w/o MFR (Ours)  & 3D R18 & 64 & 36.6 & 63.7 \\
 		\midrule
		TAM \cite{MoreIsLess19}                                   & R50 & 8 & 49.6 & 78.5 \\
		TAM+{\bf \shortname{}} w/o MFR (Ours)  & R50 & 8 & 50.1 & 79.3 \\
 		\midrule
		TSM  \cite{lin2019tsm}                                  & R50 & 8 & 46.2 & 74.3 \\
		TSM+{\bf \shortname{}} w/o MFR (Ours)  & R50 & 8 & 47.2 & 75.9 \\
		\bottomrule
	\end{tabular}
	\vspace{.1cm}
	\caption{
	{\bf Evaluating the generality of the proposed \shortname{} video attention on three different baseline action models on Something-Something-V1 \cite{goyal2017something}}.
	Setting: using the center crop and 1 clip per video in test.
	R18/50=ResNet-18/50.
	}
	\label{table:supSSV1}
\end{table}

In the main paper, as a showcase we have introduced our proposed \shortname{} module in the state-of-the-art TSM network \cite{lin2019tsm} (the baseline action model),
and conducted extensive evaluations and analysis 
on several action benchmarks.
In this supplementary, we focus on the generality analysis of 
\shortname{} video attention across different baseline action models.

\keypoint{Setting}
We use the Something-Something-V1 dataset \cite{goyal2017something}.
To examine the pure effect of \shortname{},
we exclude the proposed {\em Mature Feature guided Regularization} (MFR)
in model training.
Apart from TSM, we further evaluate
the widely adopted I3D \cite{carreira2017quo}
and the very recent TAM \cite{MoreIsLess19}
as the baseline models, separately.
\shortname{} is evaluated in the same training and testing setup as any baseline model for
fair comparison and accurate analysis.

\keypoint{Results}
The results are shown in Table~\ref{table:supSSV1}.
It is clear that \shortname{}
consistently brings about accuracy performance boost to very different baseline architectures.
This suggests strong generality of
the proposed \shortname{} attention
in improving existing action models.
It is also worth mentioning that
further applying our MFR strategy to model training, we would expect more significant performance margins
(cf. Table 5 in the main paper).

\end{document}